\documentclass{article}

\usepackage{arxiv}

\usepackage[utf8]{inputenc} % allow utf-8 input
\usepackage[T1]{fontenc}    % use 8-bit T1 fonts
\usepackage{hyperref}       % hyperlinks
\usepackage{url}            % simple URL typesetting
\usepackage{booktabs}       % professional-quality tables
\usepackage{amsfonts}       % blackboard math symbols
\usepackage{nicefrac}       % compact symbols for 1/2, etc.
\usepackage{microtype}      % microtypography
\usepackage{lipsum}		% Can be removed after putting your text content
\usepackage{graphicx}
\usepackage[numbers]{natbib} % Use numeric references
\usepackage{doi}

\title{Handheld Video Document Scanning: A Robust On-Device Model for Multi-Page Document Scanning}

\author{ \href{https://orcid.org/0009-0002-2051-0043}{\includegraphics[scale=0.06]{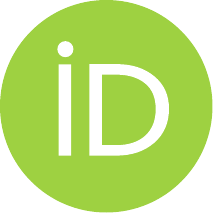}\hspace{1mm}Curtis Wigington} \\
	Adobe Research - Document Intelligence Lab \\
	College Park, Maryland \\
	\texttt{wigingto@adobe.com}
}

\date{}

\renewcommand{\headeright}{}
\renewcommand{\undertitle}{}
\renewcommand{\shorttitle}{Handheld Video Document Scanning}

\hypersetup{
pdftitle={Handheld Video Document Scanning},
pdfauthor={Curtis Wigington},
}

\begin{document}
\maketitle

\begin{figure*}[ht]
    \centering
    \includegraphics[width=\textwidth]{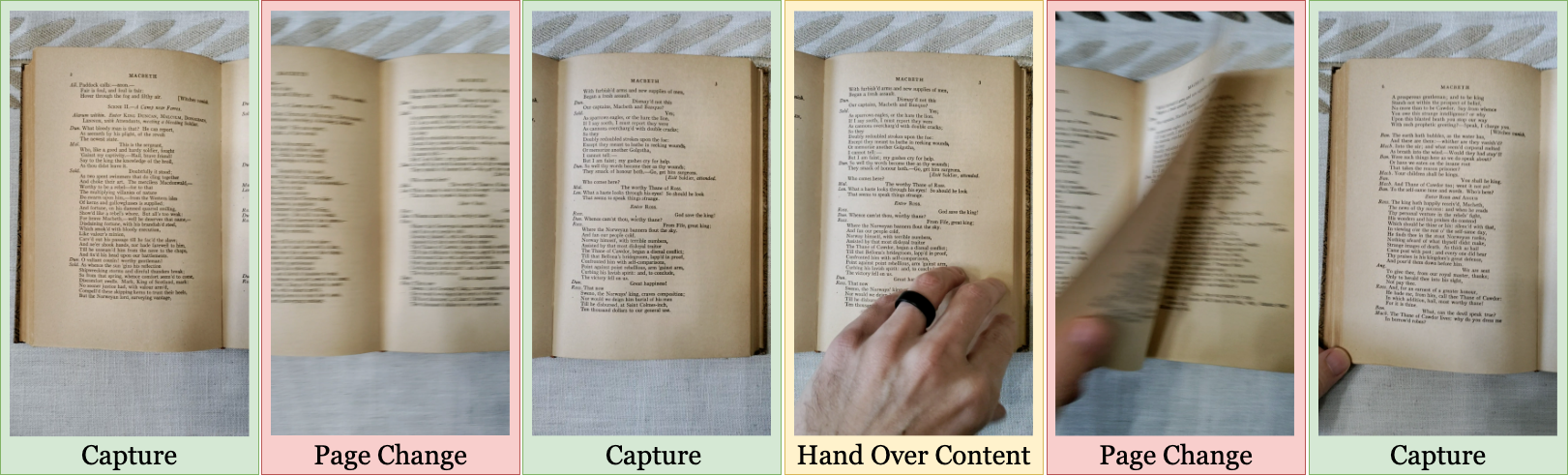} % Replace with your image file
    \caption{Sample video frames during an automatic video capture and model predictions. After capturing the left page, the user pans to the right page. The model detects the change and waits for a frame to detect problems such as the user's hand over the page content. As the user turns the page, the model detects the change and waits to capture.}
    \label{fig:teaser}
\end{figure*}

\begin{abstract}
Document capture applications on smartphones have emerged as popular tools for digitizing documents.
For many individuals, capturing documents with their smartphones is more convenient than using dedicated photocopiers or scanners, even if the quality of digitization is lower. 
However, using a smartphone for digitization can become excessively time-consuming and tedious when a user needs to digitize a document with multiple pages.

In this work, we propose a novel approach to automatically scan multi-page documents from a video stream as the user turns through the pages of the document.
Unlike previous methods that required constrained settings such as mounting the phone on a tripod, our technique is designed to allow the user to hold the phone in their hand.
Our technique is trained to be robust to the motion and instability inherent in handheld scanning.
Our primary contributions in this work include: (1) an efficient, on-device deep learning model that is accurate and robust for handheld scanning, (2) a novel data collection and annotation technique for video document scanning, and (3) state-of-the-art results on the PUCIT page turn dataset.
\end{abstract}

\section{Introduction}

The use of smartphone cameras for document digitization, whether through the default camera or a dedicated app, has become commonplace. While dedicated scanners can achieve higher quality scans and specialized feed-forward scanners are significantly faster, the convenience and availability of using a smartphone are often necessary, particularly for remote workers.

Much research has been conducted to bridge the gap between dedicated document scanners and smartphone document scanning, including: improved OCR and Handwriting Recognition for camera-captured documents\cite{hw}, automatic document dewarping\cite{dewarp2}\cite{dewarp}, robust document binarization\cite{bin}, shadow removal\cite{shadow}, automatically finding the corners and edges of a document page\cite{smartdoc}, and automatic capture of multiple document pages from a video stream\cite{ocr_video}\cite{pucit}.

From a practical perspective, another important consideration for smartphone document digitization is the need for efficient, on-device machine learning models. As recent research increasingly makes use of deep learning, many models are too computationally expensive to run on smartphones. While hosting large models behind an API may work in some cases, server costs, network availability, and latency may make such models impractical.

In this work, we propose a novel approach to automatically scan multi-page documents from a video stream as the user turns through the pages of a document.
The general process is shown in Figure~\ref{fig:teaser} where our system classifies video frames as: capture frames, page change frames, and problematic frames (in this case a hand obscuring content).
Our system is implemented using efficient on-device models so the classification is real-time, giving the user live feedback during the capture.
Our primary contributions in this work are:
\begin{enumerate}
    \item An efficient on-device, cascading model for video capture as a live-streaming experience as opposed to post-processing of an entire video.
    \item A novel approach to efficiently collect and annotate document scanning videos.
    \item State-of-the-art results on the PUCIT Page Turn dataset.
\end{enumerate}

\section{Related Work}

Few works have explored automatically capturing document pages from a video stream.
Chakraborty et al.\cite{ocr_video} proposed a technique to perform OCR from the video stream of turning through the pages of a book.
The first stage of their approach classifies frames as flipping or non-flipping pages.
An SVM model is used to perform this classification using geometrical features.
After classifying pages as flipping or non-flipping, and a noise correction process, the middle frame from each block of non-flipping frames is selected as the ideal frame.
This work focuses on the downstream task of OCR and as a result, evaluates based on the accuracy of OCR systems on the extracted frames.
The technique was tested on videos captured from a fixed camera with three books.

Tariq and Khan\cite{pucit} similarly propose a method for automatic video-based document capture.
They collected and publicly released a dataset of 37 videos captured from a mounted smartphone except for one video where the smartphone was handheld.
Their approach consists of three main steps.

First, they detect frames that represent a page turn event.
The technique for this step was based on the observation that there is significantly more motion while the page is being turned; in particular, because the phone is mounted, there should be very little movement otherwise.
The second step identifies if a hand is present in the remaining frames.
The presence of hands in the image is based on a simple skin-color-based hand detector.
The middle frame of any blocks that are determined to not contain a hand is selected as a candidate frame.
Finally, in the third step, a SIFT\cite{sift}-based matching approach removes any duplicate images.

While these techniques showed promising results, the reliance on a steady, mounted smartphone is not practical or convenient for the average user of a document scanning app.
Additionally, these approaches do not consider the real-time streaming use case where the user is interacting with the algorithm live.
Post-processing a video after the fact is less practical than a live, real-time interaction because the user can ensure each frame is captured correctly.
If captured with a video in advance and then post-processed, there may be scenarios where none of the frames for a particular page were of sufficiently high quality for capture.
In such cases, it would require the user to find and recapture all problematic pages, negating any speed benefits from the automatic capture.
By having a live interaction, the user can immediately notice any errors.

The SmartDoc-QA\cite{smartdocqa} dataset was created to evaluate quality assessment methods for document images.
While not a video scanning method, evaluating quality-related attributes is an important step in our system.
The documents in this dataset were captured under a variety of conditions with different levels of lighting, blur, and perspective angles.
Another document quality assessment work\cite{qadataset} collected 19,943 camera-captured document images and labeled them on a scale of 0-10.
While the images from these datasets are useful for quality assessment, the types of issues that arise in document video scans are more diverse.
For example, frames where the page is in motion or a hand is covering text content.

While there is limited work on video-scanning for documents, other processing of camera-captured documents has received a lot of attention.
These include automatic corner and boundary detection\cite{smartdoc}, document dewarping\cite{dewarp}\cite{dewarp2}, shadow removal\cite{shadow}, and document enhancement or binarization\cite{bin}.
While these improve the quality of the capture, they are generally applied after the capture and not during capture.
As is done in many document scanning applications, automatic corner and boundary detection can be shown live during capture.

\section{Data Collection and Annotation}

First, we will describe our process for data collection and annotation.
Data collection for document camera capture use cases is difficult due to the general unavailability of such documents on the web.
While many people capture documents with their phones, usually the documents are not distributed until after they have been post-processed by cropping to the document boundary, dewarped, and undergone additional cleaning or enhancement.
As such, the original, raw captures to use for training are generally not available through public web sources.
Many prior works have overcome this limitation by generating synthetic data, adding noise and deformations to clean documents\cite{syn_dewarp}.
Then, any available real data is limited for fine-tuning or evaluation.

For our use case of automatic video scanning, there is even less data publicly available because it would be very unusual for someone to record the process of capturing a document page.
Furthermore, it is more difficult to generate realistic synthetic data compared to other document scanning use cases.
To the best of our knowledge, the PUCIT Page Turn Dataset\cite{pucit} is the only publicly available dataset for document video capture.
This dataset consists of 37 videos where a phone is mounted over the document or book to be captured as a person flips through the pages.
There is one example where the camera is held by the person scanning the document, leading to a less stable and more challenging video.
Additionally, the authors diversified the dataset by including books, stapled documents, and unstapled/unbound pages.
In addition to releasing the dataset, Tariq and Khan\cite{pucit} proposed a video scanning technique that did not require any machine learning.
As a result, this dataset worked well for evaluating their method but would likely be too small for training modern deep learning-based approaches.

Due to the lack of available video capture data, data collection and annotation were necessary to train a deep learning-based solution.
We will outline the considerations and methods we used to collect our dataset.

\subsection{Data Collection}

We wanted our document video scanning model to be robust to variation due to differences in cameras, document types, and capture styles.
This was best accomplished by using Mechanical Turk (MTurk) to source our video files.
Our task on MTurk gave the workers the following requirements:

\begin{itemize}
  \item Use their default or preferred video recording app to record a video of themselves turning through the pages of a book.
  \item Use a notebook or book they already have that does not contain any personal information.
  \item Capture only 10 pages.
  \item Each user was allowed to complete the task a maximum of 10 times. For each task, they should use a different notebook or book. This helped ensure a wide diversity in the videos.
  \item They were assigned one of four capture methods: (1) right page only, (2) left page only, (3) both pages at the same time, (4) both pages but panning the phone between the left and right pages. Example videos of the assigned capture method were provided.
\end{itemize}

We employed both manual and automatic methods for reviewing the videos and removed videos that were spam or were not completed according to the instructions.
Spam submissions included re-uploading the same video for multiple task, uploading non-video files, and stock videos of irrelevant content.
In most cases, these problematic videos were automatically detected.
We manually reviewed all the videos to ensure the tasks were done correctly.
We chose not to reject videos for performing the incorrect capture method or for capturing too few or too many pages as we felt these did not lower the quality of the dataset.
We found examples where a worker submitted many times under different MTurk accounts to bypass the 10-jobs-per-person requirement.
These videos were removed from the set to prevent over-representing a particular person in the dataset.
After some videos were rejected, workers corrected their mistakes, and we refunded the rejected tasks.

The final dataset consists of 632 videos from 144 unique individuals.
This represents a wide variety of camera resolutions, quality, video frame rates, page turning speeds, skin tones (detecting the presence of a user's hand is important for knowing when to capture), and page content.

\subsection{Data Annotation}

There are two key events that occur in document video scanning: Page Change Events (PCE) and Capture Events (CapE).
Both of these events are difficult to annotate for training and evaluation due to their subjective nature.
A PCE represents a range in time while the user transitions from one page to another, either by turning the page or panning over to a new page.
In this case, the exact moment when the PCE begins or ends is subjective to the annotator.

A CapE represents a range in time when the document is ready to be captured, either by using the video frame directly or by triggering a capture sequence to take a photo.
This is even more subjective as multiple factors could influence whether the page is ready for capture, such as shadows, the page being partly out of frame, or the user's hand covering the page.
The severity of each of these issues could play a role in determining if a CapE should be triggered.
Such ambiguous tasks are difficult to annotate through crowdwork platforms such as MTurk.
In the following sections, we describe how we annotated both PCE and CapE for our video dataset.

\subsection{Page Change Events}

PCEs are important to delimit between the pages that need to be captured in the video stream.
Determining the exact moment when a PCE starts or ends is not critical for our use case as long as it delimits the CapEs.

Since we only need PCEs as delimiters, for training, we annotate the PCEs as a single frame when the PCE is approximately 50\% complete.
We used MTurk for these annotations as well, and for the task, we instructed the workers to label the frame where the page is 50\% of the way turned or the phone has panned 50\% of the distance to the new pages.
While there is still some level of ambiguity, it is easy to describe and completed very quickly using our interface.

\begin{figure}[h]
  \centering
  \fbox{\includegraphics[width=0.50\linewidth]{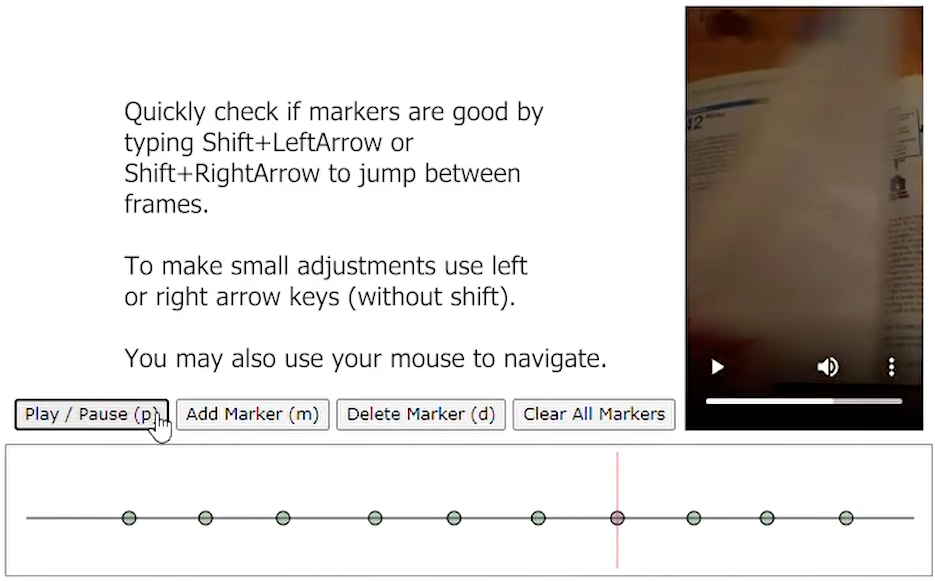}}
  \caption{Our Mechanical Turk annotation interface. Most videos could be annotated in a single playthrough of the video by clicking the "m" hotkey to mark page changes while the video was playing. If needed, the annotations could be quickly adjusted using hotkeys.}
  \label{gui}
\end{figure}

Figure \ref{gui} shows our annotation interface where the video frame is labeled as a PCE because the page turn is 50\% complete.
In this interface, after the worker starts playing the video, they press "m" on the keyboard to mark the middle of the PCE.
Using the arrow keys, the user could quickly jump between marked PCEs in the video, fine-tune the placement of PCEs, or add any missed PCEs.
We found that among the general pool of MTurk workers, there were varying degrees of quality in the work.
After running our initial pilot, we limited the full job to a few workers who performed well, resulting in much higher quality annotations compared to the pilot.

\subsection{Capture Events}

Determining whether a particular frame qualifies as a CapE can depend on a variety of factors specific to the use case and the robustness of the post-processing that follows.
Because the MTurk workers captured the video by simply recording as if they were scanning, without aiming for final scans, some videos do not have a single valid CapE or only have a CapE for a subset of the pages.
This is problematic because even in an ideal setting, CapE frames would still represent a minority of the frames.
As a result there is a significant class imbalance between between CapE frames and non-CapE frames.

For this reason, we labeled capture-related attributes (CapE Attributes) for frames instead of labeling CapE directly.
This way, even in videos where there is no valid CapE, the model can learn to recognize various attributes.
The combination of these attributes can then be used to determine a CapE.

\begin{figure}[t]
  \centering
  \includegraphics[width=\linewidth]{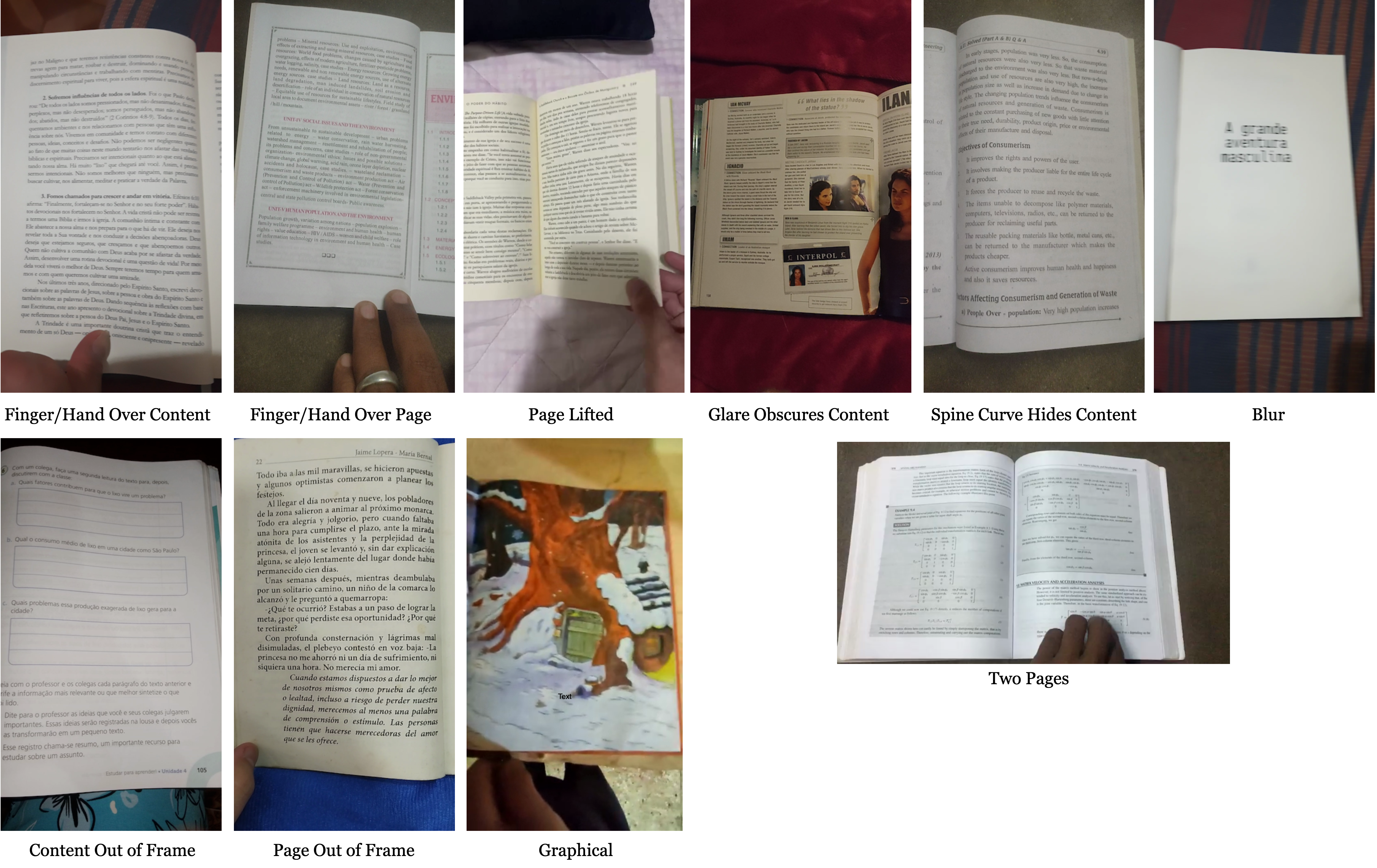}
  \caption{Example frames for each of the CapE attributes labeled in the dataset.}
  \label{fig:attr}
\end{figure}

We labeled the following attributes (an example of each attribute is shown in Figure~\ref{fig:attr}):

\begin{itemize}
    \item Hand or figure covering content
    \item Hand or figure covering page, but not covering content
    \item Page content out of frame
    \item Page out of frame, but no content out of frame
    \item Out of focus or motion blur
    \item Content obscured by glare
    \item Page lifted (slightly different than PCE)
    \item Two pages
    \item Graphical
\end{itemize}

For the hand or finger attributes and the out-of-frame attributes, we distinguish between whether or not content is lost.
This is because a user may need to use their finger to keep a book open, and some amount of covering the page may be acceptable.
Likewise, for the out-of-frame attributes, someone may take the picture from close up such that the margins are slightly clipped.
This may not be an issue if no content is lost.

Regarding the page lifted attribute, this is similar to a PCE except it can represent any time the user is in the process of turning the page and would not include when the user is panning the phone to a different page.
While "Two pages" and "Graphical" are not quality-related attributes, they are still useful.
The two-page prediction may be useful for downstream tasks such as applying extra post-processing to split the image into two pages.
We found that graphical pages were frequently misclassified because there is a fairly strong bias towards plain text pages in the dataset.
As a result, by specifically annotating graphical pages, the model was better able to handle them.
In addition to the attributes, we also annotated CapE by whether any content on the page was lost, obscured, or blurred and our personal judgment on an other factors.

These attributes were annotated exclusively by the authors.
Since there are over 1 million total frames in this dataset, it would not be practical to label all frames for all attributes.
We employed an active training loop where an initial, small set of positive and negative samples for each attribute was labeled.
Then, a CNN-based model was trained on this set.
New samples were selected based on the CNN's predictions, where the prediction activations were near 1, near 0, and near 0.5.
This way, we were able to quickly correct false positives, false negatives, and cases where the model was uncertain.
This cycle was repeated for each attribute, with some attributes requiring more annotation than others.

\section{Model}

At a high level, the proposed system predicts the start and end of PCEs and frame-wise CapEs.
We envision the video scanning user experience as a live, interactive process rather than background processing of a pre-recorded video.
Therefore, our system needs to run in real-time, and the model cannot review the entire video in advance; decisions must be based on the current or previous frames without considering future frames.

Additionally, practical considerations due to limitations of smartphones need to be taken into account.
Due to resource contention on the phone, frame rates may fluctuate and frames may drop, especially if additional post-processing is performed on captured pages.
Also, processing frames at a high enough resolution so that any frame could be saved as a high-quality capture may not be feasible while maintaining a sufficiently high frame rate.
In such cases, the video stream is configured at a lower resolution, and when a CapE is detected, the camera capture is triggered to take a photo at high resolution.
Since this process is not instantaneous, practically the system only gets one opportunity to capture the CapE per PCE.

Considering these practical limitations of real-time models, the following sections describe the models and architectures used.

\subsection{Architecture}

Our system makes use of two cascading models: a lightweight CNN-LSTM model referred to as the Page Change Network (PCN), and a MobileNet-V2 model\cite{mobilenetv2}, referred to as the Capture Network (CapN).
The PCN is designed to run faster than our target frames per second (FPS) so that it can "catch up" in case of delays due to resource contention or after the CapN has run.

\begin{figure}[h]
  \centering
  \includegraphics[width=0.5\linewidth]{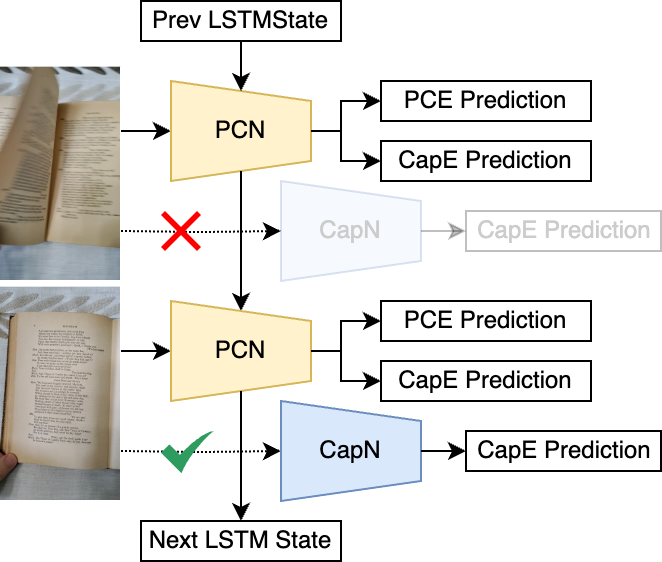}
  \caption{Overview of our proposed system: On the first frame, because the frame is in the middle of a PCE, the CapN is not run. On the second frame, the PCN predicts it is not a PCE and likely a CapE, so the slower but more accurate CapN runs.}
  \label{overview}
\end{figure}

The PCN has two outputs: it predicts PCEs and CapE attributes for every frame.
Because CapE attributes require a detailed understanding of the frame, the low-resolution, lightweight PCN does not perform this task as accurately; its purpose is simply to filter out frames that are obviously not CapEs to avoid running the more computationally expensive CapN.
The CapN runs at a higher resolution and is a larger network that produces more accurate predictions, but it only runs after a PCE has occurred and passed initial filtering by the PCN.
Figure~\ref{overview} shows an overview of our proposed system where the PCN runs on every frame, but CapN only runs when the PCN predicts a CapE with sufficient confidence.

\subsubsection{Page Change Network CNN-LSTM}

The CNN part of the CNN-LSTM resembles a MobileNetV2 network but is reduced in size, inspired by efficient CNN-LSTM networks used for OCR and Handwriting recognition \cite{hw}.
The input shape is 64x64x1.
Many smartphones default to providing frames in the YUV format, where chroma subsampling results in the Y component having a different shape from the UV components.
To avoid the overhead of converting YUV to RGB and to circumvent compatibility issues arising from varying chroma subsampling across devices, we use grayscale images as input for the PCN.
This approach is also computationally cheaper on the first convolution layer.

The CNN is a reduced version of MobileNetV2, starting with an identical first layer and utilizing only the first nine bottleneck layers before truncating the network.
Global max pooling is applied to the output of the truncated CNN, and the pooled features are then passed to the LSTM, which consists of two layers, each with a hidden size of 64.
Finally, the two output heads—one for PCEs and the other for CapE attributes—are implemented as two fully connected layers on the output of the LSTM.

We propose a variation of the CNN-LSTM, where multiple frames are processed simultaneously.
Instead of passing a single frame into the CNN-LSTM, $N$ frames are processed in one step, thereby reducing the number of forward passes by a factor of $N$.
A minimal number of additional parameters are introduced to the network: the number of input channels in the first convolutional layer is expanded from 1 to $N$, and the number of output layers increases from 2 to 2x$N$.
With this variation, $N$ frames are accumulated and processed together through the network.
By reducing the number of forward passes, slower smartphones can achieve a higher FPS.
It is worth noting that if a PCE or CapE occurs in the first frame, the user will not be informed until after the $N$th frame.
For low values of $N$, this should not have a notable impact on the end-user experience.

\subsubsection{Capture Network CNN}

For the CapN, we use a standard MobileNetV2 model to predict the final CapE attribute predictions.
The default input shape of MobileNetV2 is 224x224, which we increase to 320x320 to help the model capture more detail.

\subsection{Model Training}

The CapN network is trained first.
It follows a standard classification training procedure using the Binary Cross Entropy (BCE) loss for each CapE attribute.
Each frame is not guaranteed to be labeled with all CapE attributes, so the BCE loss is masked based on the available labels.

The CapE part of the PCN network is distilled from the CapN.
We use the fully trained CapN to run inference on all frames in the dataset and store the output probabilities for each attribute.
During the training of the PCN, these soft labels are used to train the CapE outputs of the PCN.

Because our dataset was collected from a large variety of devices and workers, there is significant variation in resolution, duration, frame rate, and aspect ratios.
During training, we try to normalize these values where possible.
We resize the images to 64x64 without distorting the aspect ratio by using padding.
We resample the videos to a target frame rate by either removing or repeating frames.
To allow for the LSTM to learn from sufficient context while still maintaining a sufficiently large batch size, we randomly select a window of 128 frames from the video after it has been adjusted to the target frame rate.

Additionally, We employ various augmentations to the data during training to help the model better generalize. The videos in our dataset may not be rotated correctly because of how the various models of phones determine which way is up.
For most scans, the phone was held parallel with the ground, making it more difficult for the phone to correctly determine the rotation.
While we could correct this in the training set, the same issue would exist for the end users of a scanning app.
Therefore, we augment our training with random 90-degree rotations and vertical and horizontal flips.
Additionally, we randomly augment the target FPS between 20 and 40 to help account for inconsistencies in frame rate during real-time processing.

\begin{figure}[h]
  \centering
  \includegraphics[width=0.5\linewidth]{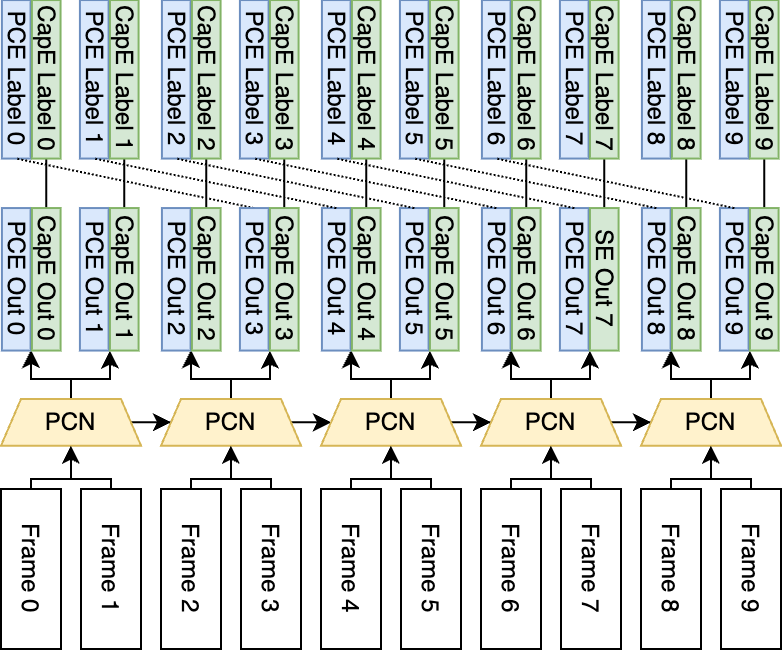}
  \caption{Visualization of look-ahead frames and processing multiple frames as input for the PCN. The CapE output uses the label of the current frame while the PCE uses the label from three frames in the past (represented with the dotted lines).}
  \label{fig:pce_training}

\end{figure}

It can be difficult for the model to know when the user is 50\% through completing a page change because they may speed up or hesitate and slow down.
For this reason, we propose to add look-ahead frames (LAFs) where the model predicts the PCE for a predefined number of LAFs in the past.
In this way, the model has the additional context of the frames that follow while only adding minimal lag to the user experience.
With a LAF of 5 frames at 30 FPS, the user experience would be delayed by 166 ms, which is likely acceptable and probably not noticeable to the user.
This is particularly true because the frames immediately following the PCE are likely not CapEs.
Note that LAF only applies to the PCE and not the CapEs as the CapEs is best determined based on only the current frame.
Figure~\ref{fig:pce_training} shows a visualization of the proposed LAFs.

The training dataset is heavily biased in that the majority of the frames are not PCEs.
Training on the PCE labels alone results in a model that rarely predicts any PCE, particularly because there is some level of ambiguity in selecting the frame at the exact 50\% mark.
To address this issue, we propose two approaches: (1) we pad the PCE labels by 4 on each side, increasing the number of positive samples, and (2) we increase the bias on the loss function to more heavily weight positive samples.

\subsection{Inference}

As previously described, we use a cascading approach where the PCN runs on every frame, making both PCE predictions and CapE attribute predictions. The CapN is only run on frames that are not PCEs and pass the threshold for the initial CapE attribute prediction made by the PCN.

From a practical perspective, many phones cannot stream video frames at a high resolution while maintaining a high enough FPS. Some of the computational cost comes from downsampling the frames to the correct size for the model. For this reason, the mobile operating systems allow for the frames to be streamed at lower resolution, and then a capture process can be triggered to take a high-resolution photo. Due to the extra overhead of taking a high-resolution photo, this should not be performed multiple times per page unless absolutely necessary. 

This motivates two types of post-processing that we will evaluate: (1) MultiCap, where any frame that gets a better CapE score can replace the previous best CapE frame, and (2) OneCap, where after each PCE, only one CapE score can be selected. For OneCap, if a frame is not selected and no better frames come along, the system cannot backtrack, and likewise, if a frame is selected and a better frame comes along, it cannot update to the new frame. MultiCap represents an ideal scenario, maybe on a higher-end phone, where all the video frames can be processed at full resolution, and OneCap represents the common scenario where frames need to be processed at a lower resolution.

The system has four hyperparameters applied to the output of the models: high PCE threshold, low PCE threshold, CapE filter threshold, and CapE threshold. The high PCE threshold is used to determine the start of a PCE, and the low PCE threshold determines the end of a PCE event. The CapE filter threshold is applied to the CapE output of the PCN to determine when to run the more expensive CapN. The CapE threshold is used to determine when a CapE event occurs and is applied to the output of CapN.

\begin{figure}[h]

  \centering
  \includegraphics[width=0.5\linewidth]{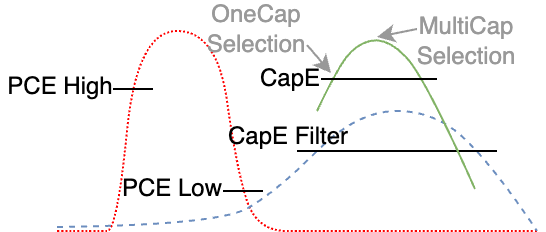}
  \caption{Visualization of the post-processing thresholds. The red dotted line and blue dashed lines are the PCE and CapE predictions from the PCN respectively. The green solid line is the CapE prediction from the CapN. The threshold values are indicated by the horizontal lines.}
  \label{fig:postprocessing}
\end{figure}

Figure~\ref{fig:postprocessing} shows a visualization of a typical page change and capture where the y-axis represents the output probability and the x-axis represents time in the video. The red dotted line represents the page change prediction; the PCE high and low thresholds determine the start and end of the PCE. The blue dotted line represents the PCN prediction for CapE, which is less accurate. The green line represents the CapN prediction for CapE. The output for the CapN is only shown for a portion of the timeline because it is only run once the network has passed the CapE filter threshold.

\section{Results}

In this section, we present our results using the publicly available PUCIT page turn dataset.
The dataset consists of 37 videos for evaluating automatic page turn capture.
The authors of the dataset annotated ranges of frames that marked acceptable CapEs.
In the original paper, it is reported that the videos were recorded at 1080p, while the videos available on the web page are of lower resolution (272x480).
As our networks do not run at very high resolution, this is not an issue, but we will not be able to run a similar SIFT-based duplication removal approach presented in their paper.

\begin{figure}[h]
  \centering
  \includegraphics[width=\linewidth]{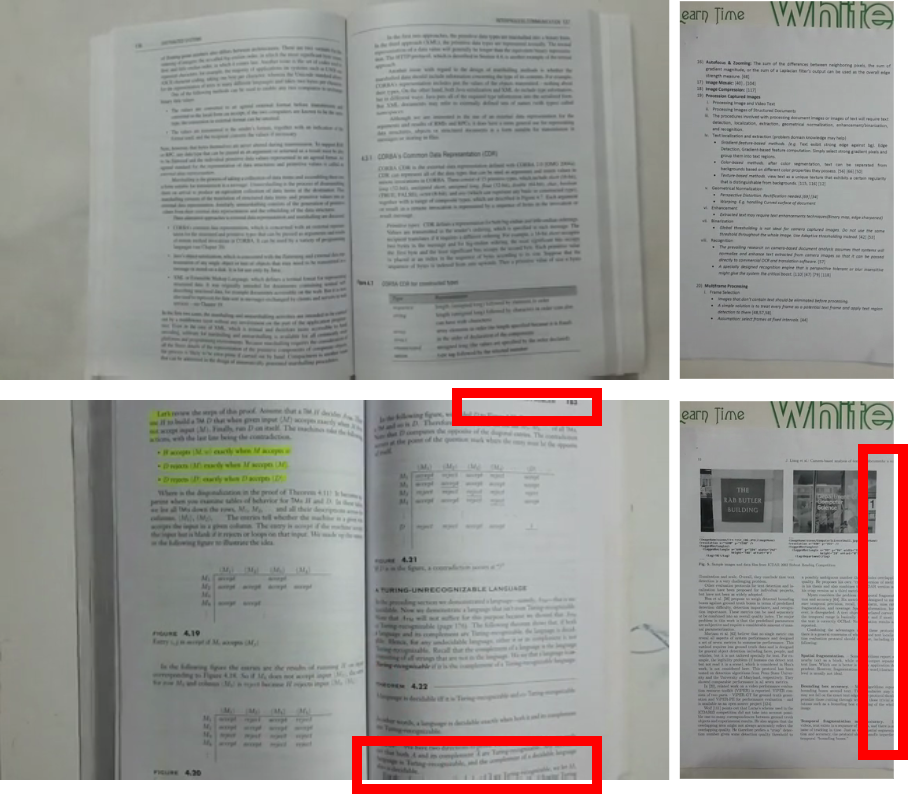}
  \caption{Samples from the PUCIT Page Turn dataset. The top row shows samples in the dataset that match our criteria for capture. The bottom row shows positively labeled samples that our model would reject due to out-of-frame content. Red boxes highlight the out-of-frame content.}
  \label{fig:pucit_clip}

\end{figure}

Additionally, the authors of the PUCIT dataset used different criteria than we did for what was considered an acceptable CapE.
Most notably, text and content (such as the header of the page) is sometimes clipped off the top and bottom of the frame, which would result in a low CapE prediction for our model because we felt frames with text and content loss did not make for a valid capture.
Figure~\ref{fig:pucit_clip} shows examples where text is clipped off and is labeled as a valid frame to capture in the dataset.

% CONTINUE correction
To accommodate this, our model needs to be tuned to this dataset.
As this dataset only contains 37 videos and no training and test splits have been proposed, we wanted to avoid training our deep model on the dataset.
Instead, we perform 5-fold cross-validation to train 5 Linear Regression models with the CapE attributes as input to predict the CapE frames.
All results reported on this dataset used this approach, and the Linear Regression weights used on any particular video came from the fold that was held out in the cross-validation.
We feel this is a fair way to report on this dataset, as the Linear Regression model only has 10 parameters, making overfitting unlikely. Additionally, we never train directly on any test video for which we report results.
We believe this simple adaptation highlights the benefit of the attribute-based prediction approach we use.

We validate our method in the OneSnap and MultiSnap scenarios.
In Table~\ref{tab:sota}, we show how our model compares with \cite{pucit}.
Our model outperforms the existing approach in both the OneSnap and MultiSnap scenarios.
This is notable because both OneSnap and MultiSnap consider the video as a stream where previous frames cannot be processed in the context of future frames, making it a more difficult setting.
Additionally, in \cite{pucit}, extra experiments were conducted on one video sample where the phone was handheld instead of mounted, as it was a more challenging case.
Our model achieved an F1 score of 1.0 on this sample; since all of our training data came from phones that were handheld, our model was robust to the extra noise introduced in this example.

\begin{table}
  \centering
  \caption{Comparison with State-of-the-Art}
  \begin{tabular}{rlll}
    \toprule
    Method & P & R & F1 \\
    \midrule
    Tariq and Khan\cite{pucit} & 0.92 & 0.90 & 0.91 \\
    Ours - OneSnap & 0.94 & 0.94 & 0.94 \\
    Ours - MultiSnap & 0.97 & 0.97 & 0.97 \\
  \bottomrule
  \label{tab:sota}
\end{tabular}
\end{table}

\subsection{Ablation Study}

\subsubsection{Capture Network Threshold}

In our proposed method, we use a cascading model where the CapN only runs after the output of the PCN reaches a specific threshold, as shown by the green line in Figure~\ref{fig:postprocessing}.
This threshold can be tuned to adjust how frequently the CapN runs.
Table~\ref{tab:sn_balance} shows the percentage of frames on which the CapN runs and the F1 score for varying threshold values for both the OneCap and MultiCap scenarios.
In the MultiCap scenario, the model runs more often than in the OneCap scenario because the CapN stops running after capturing the first frame that passes the CapE threshold.
The MultiCap model continues to run and selects the frame that achieves the best value.
Both approaches do not run during a PCE, which is why neither reaches 100\% when the threshold is 0.

The OneCap approach is the more realistic approach for user-facing applications.
Using a threshold value of 0.6, the CapN only runs on 3.7\% of the frames with no loss in the F1 score, which is significant given the reduction in computational cost.
Based on the device timing data shown in Table~\ref{tab:device_comparision}, running the CapN a low percentage of the time is critical for operating our system on slower Android devices.

\begin{table}
  \centering
  \caption{Capture Network Threshold.}
  \begin{tabular}{ccccc}
    \toprule
     & \multicolumn{2}{c}{OneCap} & \multicolumn{2}{c}{MultiCap} \\
    Threshold & Run Percent & F1 & Run Percent & F1\\
    \midrule
    0.0 & 23.0\% & 0.94 & 91.3\% & 0.97 \\
    0.2 & 6.4\% & 0.94 & 54.5\% & 0.97 \\
    0.4 & 4.8\% & 0.94 & 50.2\% & 0.97 \\
    0.6 & 3.7\% & 0.94 & 39.2\% & 0.96\\
    0.8 & 2.4\% & 0.93 & 39.2\% & 0.96 \\
  \bottomrule
  \label{tab:sn_balance}
\end{tabular}
\end{table}

\subsubsection{Multiple Frames as Input Channels}

We propose in our method to combine multiple video frames as a single input into the network in the form of multiple input channels.
All layers in the network remain the same except for the input and output layers, which, for small values of N, have a negligible effect on the inference time of a single forward pass of the network.
This results in the network running fewer times by a factor of N.

Interestingly, when evaluating just the PCN, the F1 score of the model significantly improves with additional input frames.
However, when running the full system with both the PCN and CapN, the differences between the F1 scores for different values of N are small.
This indicates that the PCN is performing similarly well on the PCE prediction, but the merged input frames help the model more accurately predict the CapE attributes.
There is a potential negative side effect of merging frames in that the response to the user could be delayed by N frames.
As a result, a value of N of two or three frames appears to be a good balance of improved efficiency and limited delays in user experience with no loss in accuracy.

\begin{table}
  \centering
  \caption{Analysis of N input images per forward pass}
  \begin{tabular}{ccccc}
    \toprule
    Method & N & P & R & F1 \\
    \midrule
    Just PCN & 1 & 0.82 & 0.80 & 0.80 \\
    Full     & 1 & 0.96 & 0.98 & 0.97 \\
    Just PCN & 2 & 0.96 & 0.95 & 0.95 \\
    Full     & 2 & 0.97 & 0.97 & 0.97 \\
    Just PCN & 3 & 0.94 & 0.95 & 0.94 \\
    Full     & 3 & 0.97 & 0.98 & 0.98 \\
  \bottomrule
  \label{tab:n_input}
\end{tabular}
\end{table}

\subsection{On-device Runtime}

We evaluated the runtime of this model on three Android phones selected to represent a range of performance levels, including both high-end and older models.
We converted both models to TensorFlow Lite models and used TensorFlow's on-device benchmarking tool.
As this benchmarking tool runs with "adb" and not in a traditional app, it represents an idealistic runtime where the model is not competing with other apps or processes.
In a real app, the FPS would be slower due to other processes using resources.

Table~\ref{tab:device_comparision} shows the FPS on the three phones.
We ran the benchmark in both the CPU and GPU settings.
In the case of the PCN network, not all operators are supported by the GPU delegate, which results in the network being split into running the first part on the GPU and the second part on the CPU.
As the PCN network is already very lightweight, it is likely that the overhead from switching between the GPU and CPU causes a degradation in performance, and as a result, the GPU is generally slower for the PCN.

\begin{table}
  \centering
  \caption{On-device runtime comparison}
  \begin{tabular}{ccccc}
    \toprule
    Phone & Model & CPU FPS & GPU FPS \\
    \midrule
    Pixel 7 Pro (2022) & PCN & 557.4 & 65.7 \\
    Pixel 7 Pro (2022) & CapN  & 38.7  & 31.1 \\
    OnePlus N10 (2020) & PCN & 274.2 & 117.4 \\
    OnePlus N10 (2020) & CapN  & 17.2  & 31.3 \\
    Moto X4 (2017)     & PCN & 49.2  & 42.2 \\
    Moto X4 (2017)     & CapN  & 4.8   & 12.9 \\
  \bottomrule
  \label{tab:device_comparision}
\end{tabular}
\end{table}

The results in Table~\ref{tab:device_comparision} compute the FPS for the forward pass of the networks and do not take into account the speed improvements from processing multiple input frames in the same forward pass. 
This is particularly important for slower devices such as the Moto X4 which may not be able to maintain 30 FPS in a real app where there are other processes using computation resources.

\section{Conclusion}

In this work we presented a novel approach for scanning documents through a video stream.
An emphasis of our work was that it needed to be efficient to run on-device and needed to run as a live experience.
We proposed a novel approach to collect and annotate a video scan document dataset that enabled efficient annotation while avoiding burdening annotators with the inherently ambiguous nature of video scan labels.
We proposed a variety of techniques to train an efficient Page Change Network and a novel attribute-based output for the Capture Network.
We achieved state-of-the-art results using the proposed technique.

We focused on developing a system that could run even on slower devices.
While our model achieved state-of-the-art results, it would be interesting to apply our techniques using larger models intended for use on only higher-end devices and see how much additional improvement could be made.
Additionally, with recent interest in AR and improved sensors on phones, many phones have depth-sensors to augment the cameras and have highly-optimized tracking libraries designed to be used with AR applications.
It would be interesting to see how these newer features of phones could augment the camera for even better document scanning.

\bibliographystyle{unsrtnat}
\bibliography{references}

\begin{thebibliography}{13}
\providecommand{\natexlab}[1]{#1}
\providecommand{\url}[1]{\texttt{#1}}
\expandafter\ifx\csname urlstyle\endcsname\relax
  \providecommand{\doi}[1]{doi: #1}\else
  \providecommand{\doi}{doi: \begingroup \urlstyle{rm}\Url}\fi

\bibitem[Wigington et~al.(2017)Wigington, Stewart, Davis, Barrett, Price, and Cohen]{hw}
Curtis Wigington, Seth Stewart, Brian Davis, Bill Barrett, Brian Price, and Scott Cohen.
\newblock Data augmentation for recognition of handwritten words and lines using a cnn-lstm network.
\newblock In \emph{2017 14th IAPR International Conference on Document Analysis and Recognition (ICDAR)}, volume~01, pages 639--645, 2017.
\newblock \doi{10.1109/ICDAR.2017.110}.

\bibitem[Ramanna et~al.(2019)Ramanna, Bukhari, and Dengel]{dewarp2}
Vijaya Kumar~Bajjer Ramanna, Syed~Saqib Bukhari, and Andreas Dengel.
\newblock Document image dewarping using deep learning.
\newblock In \emph{ICPRAM}, pages 524--531, 2019.

\bibitem[Xie et~al.(2021)Xie, Yin, Zhang, and Liu]{dewarp}
Guo-Wang Xie, Fei Yin, Xu-Yao Zhang, and Cheng-Lin Liu.
\newblock Document dewarping with control points.
\newblock In Josep Llad{\'o}s, Daniel Lopresti, and Seiichi Uchida, editors, \emph{Document Analysis and Recognition -- ICDAR 2021}, pages 466--480, Cham, 2021. Springer International Publishing.
\newblock ISBN 978-3-030-86549-8.

\bibitem[Lins et~al.(2023)Lins, de~F.~Pe~Silva, Chaves, da~Silva~Barboza, Bernardino, and Simske]{bin}
Rafael~Dueire Lins, Gabriel de~F.~Pe~Silva, Gustavo~P. Chaves, Ricardo da~Silva~Barboza, Rodrigo~Barros Bernardino, and Steven~J. Simske.
\newblock Quality, space and time competition on binarizing photographed document images.
\newblock In \emph{Proceedings of the ACM Symposium on Document Engineering 2023}, DocEng '23, New York, NY, USA, 2023. Association for Computing Machinery.
\newblock ISBN 9798400700279.
\newblock \doi{10.1145/3573128.3604903}.
\newblock URL \url{https://doi.org/10.1145/3573128.3604903}.

\bibitem[Lin et~al.(2020)Lin, Chen, and Chuang]{shadow}
Yun-Hsuan Lin, Wen-Chin Chen, and Yung-Yu Chuang.
\newblock Bedsr-net: A deep shadow removal network from a single document image.
\newblock In \emph{Proceedings of the IEEE/CVF Conference on Computer Vision and Pattern Recognition (CVPR)}, June 2020.

\bibitem[Burie et~al.(2015)Burie, Chazalon, Coustaty, Eskenazi, Luqman, Mehri, Nayef, Ogier, Prum, and Rusiñol]{smartdoc}
JC. Burie, J.~Chazalon, M.~Coustaty, S.~Eskenazi, M.M. Luqman, M.~Mehri, N.~Nayef, JM. Ogier, S.~Prum, and M.~Rusiñol.
\newblock Icdar2015 competition on smartphone document capture and ocr (smartdoc).
\newblock In \emph{2015 13th International Conference on Document Analysis and Recognition (ICDAR)}, pages 1161--1165, 2015.
\newblock \doi{10.1109/ICDAR.2015.7333943}.

\bibitem[Chakraborty et~al.(2013)Chakraborty, Roy, Alvarez, and Pal]{ocr_video}
Dibyayan Chakraborty, Partha~Pratim Roy, Jose~M. Alvarez, and Umapada Pal.
\newblock Ocr from video stream of book flipping.
\newblock In \emph{2013 2nd IAPR Asian Conference on Pattern Recognition}, pages 130--134, 2013.
\newblock \doi{10.1109/ACPR.2013.24}.

\bibitem[Tariq and Khan(2017)]{pucit}
Waqas Tariq and Nazar Khan.
\newblock Click-free, video-based document capture - methodology and evaluation.
\newblock In \emph{2017 14th IAPR International Conference on Document Analysis and Recognition (ICDAR)}, volume~06, pages 21--26, 2017.
\newblock \doi{10.1109/ICDAR.2017.344}.

\bibitem[Lowe(2004)]{sift}
David~G Lowe.
\newblock Distinctive image features from scale-invariant keypoints.
\newblock \emph{International journal of computer vision}, 60:\penalty0 91--110, 2004.

\bibitem[Nayef et~al.(2015)Nayef, Luqman, Prum, Eskenazi, Chazalon, and Ogier]{smartdocqa}
Nibal Nayef, Muhammad~Muzzamil Luqman, Sophea Prum, Sebastien Eskenazi, Joseph Chazalon, and Jean-Marc Ogier.
\newblock Smartdoc-qa: A dataset for quality assessment of smartphone captured document images - single and multiple distortions.
\newblock In \emph{2015 13th International Conference on Document Analysis and Recognition (ICDAR)}, pages 1231--1235, 2015.
\newblock \doi{10.1109/ICDAR.2015.7333960}.

\bibitem[Li et~al.(2021)Li, Yang, Shen, and Wen]{qadataset}
Zhenghao Li, Cihui Yang, Qingyun Shen, and Shiping Wen.
\newblock A document image dataset for quality assessment.
\newblock \emph{Journal of Physics: Conference Series}, 1828:\penalty0 012033, 02 2021.
\newblock \doi{10.1088/1742-6596/1828/1/012033}.

\bibitem[Das et~al.(2019)Das, Ma, Shu, Samaras, and Shilkrot]{syn_dewarp}
Sagnik Das, Ke~Ma, Zhixin Shu, Dimitris Samaras, and Roy Shilkrot.
\newblock Dewarpnet: Single-image document unwarping with stacked 3d and 2d regression networks.
\newblock In \emph{Proceedings of International Conference on Computer Vision}, 2019.

\bibitem[Sandler et~al.(2018)Sandler, Howard, Zhu, Zhmoginov, and Chen]{mobilenetv2}
Mark Sandler, Andrew Howard, Menglong Zhu, Andrey Zhmoginov, and Liang-Chieh Chen.
\newblock Mobilenetv2: Inverted residuals and linear bottlenecks.
\newblock In \emph{Proceedings of the IEEE conference on computer vision and pattern recognition}, pages 4510--4520, 2018.

\end{thebibliography}

\end{document}